\documentclass{article}

\usepackage{microtype}
\usepackage{graphicx}
\usepackage{subcaption}
\usepackage{booktabs} 

\usepackage{hyperref}

\usepackage[preprint]{icml2026}

\usepackage{amsmath}
\usepackage{amssymb}
\usepackage{mathtools}
\usepackage{amsthm}

\usepackage{colortbl}
\usepackage{multirow}

\usepackage[capitalize,noabbrev]{cleveref}

\theoremstyle{plain}

\theoremstyle{definition}

\theoremstyle{remark}

\usepackage[most]{tcolorbox}

\usepackage{hyperref}
\usepackage{url}
\usepackage[utf8]{inputenc} 
\usepackage[T1]{fontenc}    
\usepackage{hyperref}       
\usepackage{url}            
\usepackage{booktabs}       
\usepackage{amsfonts}       
\usepackage{nicefrac}       
\usepackage{microtype}      
\usepackage{xcolor}         
\definecolor{deepgreen}{RGB}{0,100,0} 
\usepackage{xspace}
\usepackage{amsmath}
\usepackage{caption}

\usepackage{makecell}
\usepackage{multirow}
\usepackage{amsthm}
\usepackage{colortbl}
\usepackage{wrapfig}
\usepackage{caption}
\usepackage{bm}

\usepackage[normalem]{ulem}

\usepackage[most]{tcolorbox}
\usepackage{listings}
\usepackage{xcolor}  

\newtcolorbox{formalizedprompt}[1][]{%
  enhanced,            
  breakable,           
  colback=gray!5,      
  colframe=gray!80,    
  fonttitle=\bfseries, 
  coltitle=black,      
  sharp corners,       
  boxrule=0.8pt,       
  arc=2mm,             
  left=4pt, right=4pt, top=6pt, bottom=6pt, 
  width=\textwidth,    
  #1                   
}
\icmltitlerunning{Thinking with Tables: Enhancing Multi-Modal Tabular Understanding via Neuro-Symbolic Reasoning}

\begin{document}

\twocolumn[
  \icmltitle{Thinking with Tables: Enhancing Multi-Modal Tabular Understanding via Neuro-Symbolic Reasoning}



  \icmlsetsymbol{equal}{*}

  \begin{icmlauthorlist}
    \icmlauthor{Kun-Yang Yu}{yyy,comp}
    \icmlauthor{Zhi Zhou}{yyy}
    \icmlauthor{Shi-Yu Tian}{yyy,comp}
    \icmlauthor{Xiao-Wen Yang}{yyy,comp}
    \icmlauthor{Zi-Yi Jia}{yyy,sch}
    \icmlauthor{Ming Yang}{yyy,comp}
    \icmlauthor{Zi-Jian Cheng}{yyy,sch}
    \icmlauthor{Lan-Zhe Guo}{yyy,sch}
    \icmlauthor{Yu-Feng Li}{yyy,comp}
  \end{icmlauthorlist}

  \icmlaffiliation{yyy}{State Key Laboratory of Novel Software Technology, Nanjing University, China}
  \icmlaffiliation{comp}{School of Artificial Intelligence, Nanjing University, China}
  \icmlaffiliation{sch}{School of Intelligence Science and Technology, Nanjing University, China}

  \icmlcorrespondingauthor{Lan-Zhe Guo}{guolz@lamda.nju.edu.cn}
  \icmlcorrespondingauthor{Yu-Feng Li}{liyf@nju.edu.cn}

  \icmlkeywords{Machine Learning, ICML}

  \vskip 0.3in
]



\printAffiliationsAndNotice{}  

\begin{abstract}
Multimodal Large Language Models (MLLMs) have demonstrated remarkable reasoning capabilities across modalities such as images and text. However, tabular data, despite being a critical real-world modality, remains relatively underexplored in multimodal learning. In this paper, we focus on the task of Tabular-Vision Multi-Modal Understanding (TVMU) and identify three core challenges: (1) high structural variability and data incompleteness in tables, (2) implicit and complex feature dependencies, and (3) significant heterogeneity in problem-solving pipelines across downstream tasks. To address these issues, we propose Thinking with Tables (TWT). TWT employs a program-aided code-based neuro-symbolic reasoning mechanism that facilitates key operations, such as information extraction and element modeling, by interacting with external environments. We evaluate TWT on eight representative datasets. Experimental results demonstrate that TWT consistently outperforms existing baselines by an average of 10\% in accuracy, achieving performance comparable to, or even surpassing, proprietary commercial SOTA LLMs on TVMU tasks. Models and codes are available at \url{https://github.com/kunyang-YU/Thinking-with-Tables}
\end{abstract}

\section{Introduction}
In recent years, MLLMs ~\cite{achiam2023gpt,Qwen2.5-VL,qwen3technicalreport,team2023gemini} have advanced rapidly. By jointly modeling heterogeneous modalities such as text and images, MLLMs have achieved remarkable success in tasks including image understanding~\cite{shen2025skywork} and visual question answering~\cite{zhang2025thyme}, demonstrating strong complex reasoning capabilities~\cite{peng2025lmm,team2024qwq,guo2025deepseek}. In real-world applications, tabular data constitute an equally common and important modality~\cite{naomi17tabular}, widely present in domains such as finance~\cite{ozbayoglu2020deep}, healthcare~\cite{ching2018opportunities}, and industrial manufacturing~\cite{HeinDTUHRS17}, where they play a crucial role in information organization, data analysis, and decision support~\cite{Salehpour2024economics}. However, existing research on MLLMs has primarily focused on modalities such as text and images, with relatively limited attention paid to multimodal understanding involving tabular data.

Tabular-Vision Multi-Modal Understanding (TVMU)~\cite{pasupat2015compositional,petfinder-adoption-prediction,alonso2025tablet} aims to extract and reason over tables from images using textual and table-associated visual information~\cite{dang2024sadl}. It primarily encompasses two core tasks: multimodal table question answering~\cite{lu2022dynamic} and multimodal table prediction~\cite{petfinder-pawpularity-score}. Code-based neuro-symbolic approaches that use symbolic code for structured logic and computation are widely regarded as promising solutions for such tasks~\cite{chen2022program, gao2023pal}. In the Table QA domain, methods represented by Chain-of-Table~\cite{wang2024chain} decompose complex queries into a sequence of predefined table operations and derive the final answer by executing these operations step by step. For Table Prediction tasks, OCTree~\cite{nam2024optimized} leverages the background knowledge of language models to assist feature engineering in the form of executable code, thereby improving model performance. 

However, these methods typically rely on complete and well-structured tables~\cite{jiang2025multimodal,jiangcompositional}, are constrained by a limited set of predefined operations~\cite{chegini2025repanda, yang2024language}, and are often designed for a single task setting.
To enhance the capability of MLLMs in TVMU tasks, a unified framework that simultaneously supports different downstream tasks is required. For example, in financial analysis~\cite{ozbayoglu2020deep}, key information in complex research reports is often scattered across structured tables, unstructured textual analyses, and trend charts. Analysts must not only predict future growth from historical tabular data but also reason over textual market dynamics to identify the causes of profit fluctuations. In such settings, tables may be incomplete or poorly formatted, and critical decision-relevant information spans multiple modalities, making reliance on tables alone insufficient. As a result, robust modeling and reasoning for TVMU in realistic multimodal environments face three major challenges: 1) structural variability and data incompleteness in real-world applications, 2) implicit and complex feature dependencies, and 3) the significant heterogeneity of task requirements.

To address the above challenges, we propose \textbf{Thinking with Tables} (TWT), \textbf{a program-aided neuro-symbolic reasoning framework} designed for Tabular-Vision Multi-Modal Understanding tasks. TWT performs code-based neuro-symbolic reasoning by modeling table understanding as an interactive reasoning process within an external tabular environment, enabling multi-step interactions for dynamic information extraction, key element modeling, and so on. This design decomposes complex table reasoning into a sequence of interpretable intermediate steps. Experiments on eight datasets covering both multimodal table question answering and multimodal table prediction tasks demonstrate consistent and significant performance improvements, validating the effectiveness and robustness of TWT in complex TVMU tasks.

The main contributions of this work are summarized as follows:

\begin{enumerate}
\item \textbf{TVMU Problem Formulation and Analysis}. We provide a systematic analysis of the TVMU task, pointing out key challenges including table structural variability, information incompleteness, complex feature dependencies, and task heterogeneity.

    \item \textbf{Noval Neuro-Symbolic Reasoning Mechanism}. We introduce \textbf{Thinking with Tables}, a novel neuro-symbolic framework that enables robust modeling of complex tabular information through a code-based interactive reasoning mechanism.

    \item \textbf{Superior Empirical Performance} Extensive experiments on diverse datasets confirm that TWT consistently outperforms baselines for both question answering and prediction tasks, validating its robustness and adaptability in handling complex table-related tasks.
\end{enumerate}

\section{Related Works}
\subsection{Multi-Modal Tabular Understanding}
Multimodal tabular understanding~\cite{jiang2025multimodal} aims to parse table structures from document images or scanned copies and to perform deep semantic reasoning by jointly leveraging textual context and table-associated visual information~\cite{gu2022xylayoutlm,kim2021donut}. It mainly involves two categories of tasks: multimodal table question answering~\cite{jiangcompositional,peng2023table} and multimodal table prediction~\cite{ye2024revisiting,jiang2025representation,hollmann2025accurate,petfinder-adoption-prediction,zhou2025fully,yuhandling,cheng2025fully}. For multimodal table question answering, many existing studies~\cite{zhuang2024structlm, zhang2025deepanalyze,tian2025tabularmath} enhance reasoning capability by constructing multiple reasoning paths or introducing chain-of-thought style mechanisms~\cite{wei2022chain, lei2025reasoning}. In contrast, methods for multimodal table prediction~\cite{jiang2024tabular}, such as TabLLM~\cite{hegselmann2023tabllm}, are largely designed for unimodal tabular or textual inputs and thus cannot effectively handle multimodal and multi-task scenarios~\cite{du2024tip,luo2025time,tian2024crosel}. Moreover, research that unifies multimodal table question answering and prediction within a single framework remains relatively limited.

\subsection{Neuro-Symbolic Reasoning}
Neuro-symbolic reasoning~\cite{yang2025neuro, manhaeve2018deepproblog} aims to combine traditional symbolic reasoning with learning-based mechanisms to improve the reliability of intelligent systems~\cite{wang2019satnet}. 
In the era of large language models, reasoning is mainly enhanced through three symbolic paradigms: \textbf{symbolic solver–aided methods}~\cite{pan2023logic} that formalize natural language and delegate reasoning to external solvers, \textbf{program-aided}~\cite{gao2023pal} methods that execute generated programs via interpreters, and \textbf{tool-aided methods}~\cite{zhang2025thyme} that coordinate multiple external tools for flexible task-specific reasoning via SFT and RL.
Existing neuro-symbolic systems have demonstrated notable success in domains such as mathematics~\cite{tian2025vcsearch,hao2025large,zhou2025lawgpt}, coding~\cite{sapkota2025vibe}, and planning~\cite{shao2024chinatravel}. Owing to the structured nature of tabular data, neuro-symbolic reasoning is particularly well-suited for table question answering and understanding, enabling more accurate, robust, and interpretable table reasoning.
\begin{figure*}[h]
    \begin{minipage}[b]{0.37\textwidth}  
        \centering
        \includegraphics[width=\linewidth]{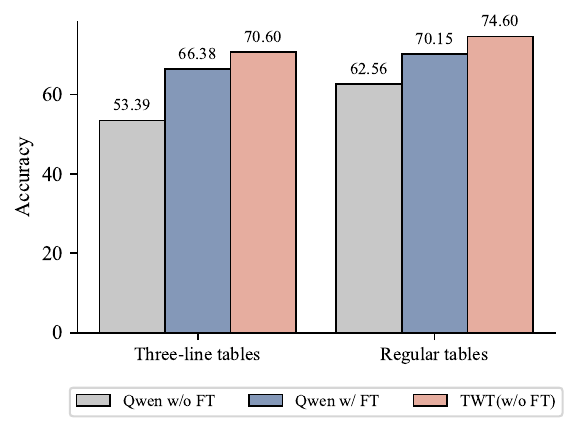}  
        \caption{Performance on TAT-QA dataset with different situations. FT means a full table message is provided.}
        \label{fig:sub1}
    \end{minipage}
    \hfill  
    \begin{minipage}[b]{0.62\textwidth}
        \centering
        \includegraphics[width=\linewidth]{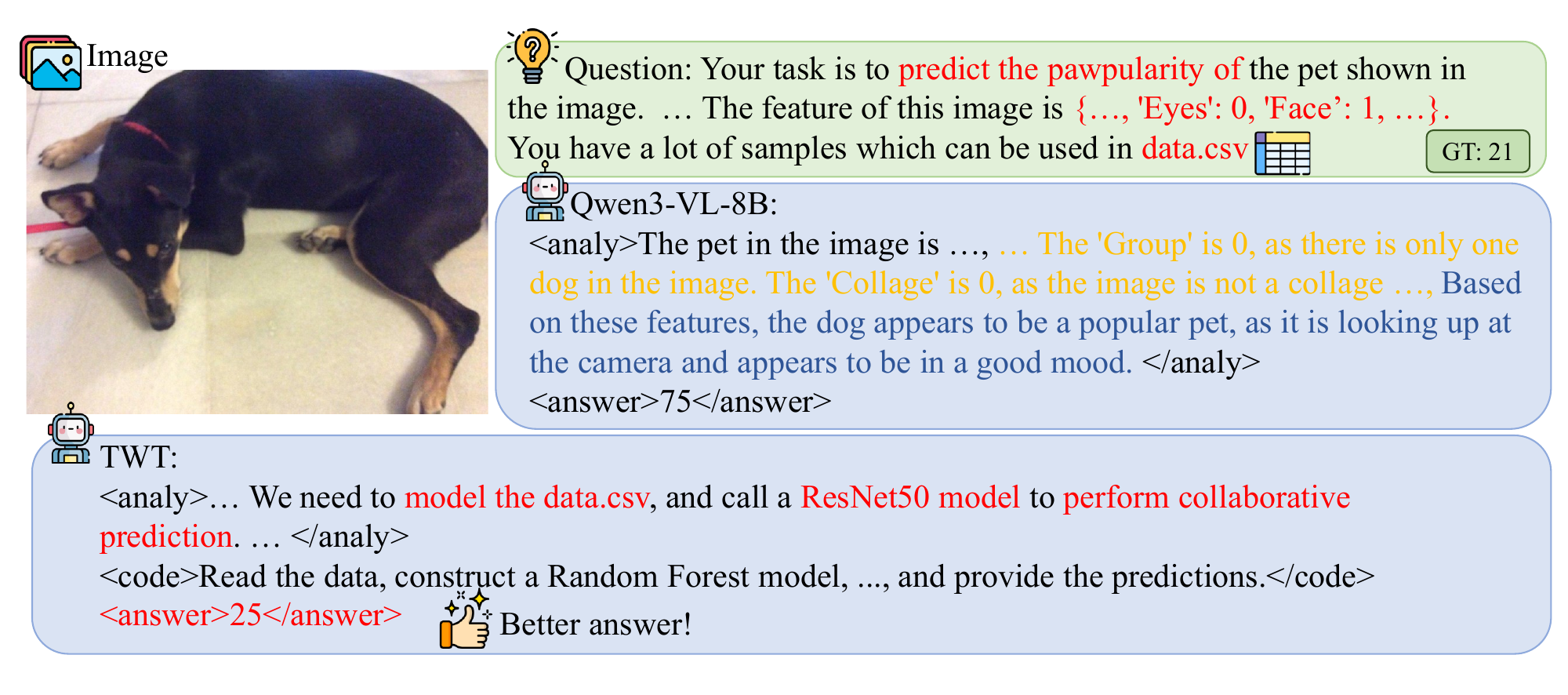}  
        \caption{Predictions from different models. Yellow highlights indicate insufficient feature modeling by existing methods, while red highlights show that our model performs more comprehensive feature modeling through code-based interactions.}
        \label{fig:sub2}
    \end{minipage}

    \label{fig:main}
\end{figure*}
\section{Problem and Analysis}
In this section, we first introduce the definition and formalization of the Tabular-Vision Multi-Modal Understanding task, and then provide a systematic analysis of the challenges associated with this task.
\subsection{Problem Definition}
Tabular-Vision Multi-Modal Understanding (TVMU) aims to parse table structures from document images or scanned copies and perform deep reasoning by jointly leveraging textual context and table-associated visual information. It primarily encompasses two core tasks: multimodal table question answering and multimodal table prediction. In this task, models do not have access to the complete table in the query. Instead, they are required to extract table structures from the visual modality, interpret table-associated images, and perform operations such as information extraction, mathematical computation, and table modeling to answer user queries.

Based on this formulation, in the TVMU task, the MLLM $\mathcal{M}$ is provided with only the table headers and question-relevant images as visual inputs $V$, while the accompanying textual context of the table and the user query $Q$ are given as textual inputs. Meanwhile, an interactive sandbox environment is introduced as the external environment $E$, which contains the raw tabular data files, image files, and potentially pre-trained models that may be required during reasoning. 

We expect the model to perform multimodal visual understanding, table modeling, and related reasoning through iterative interactions with the external environment, and ultimately generate the final answer. The generated answer is expected to be highly consistent with the ground-truth answer $y_i$, thereby improving accuracy and reliability in real-world multimodal tabular reasoning scenarios.

\[
\min_{\mathcal{M}} \sum_{i=1}^{N} \ell\bigl(\mathcal{M}(V_i, Q_i, E_i), y_i\bigr)
\]

where $\ell(\cdot)$ denotes a loss function that measures the discrepancy between the model prediction and the ground-truth answer, and $\mathcal{M}(V_i, Q_i, E_i)$ represents the prediction generated by the model given the visual input $V_i$, the query $Q_i$, and the external environment $E_i$ for the $i$-th sample. Our objective is to guide the model to effectively leverage visual information and environment interaction capabilities to accomplish table parsing, key information localization, and necessary computational reasoning. During evaluation, the overall performance is assessed based on the accuracy of the answers generated by the model under the same setting.

\subsection{Problem Analysis}

\textbf{\underline{Challenge 1}:} \textbf{Structural variability and data incompleteness in real-world applications.}
In practical applications, tables can take highly varied forms, such as three-line tables, merged-row or merged-column tables, and tables embedded within document pages rather than presented independently~\cite{zhu2021tat, chang2008bigtable}.  Moreover, in scenarios involving scanned documents, low-resolution screenshots, or tables affected by occlusion or layout degradation, numerical and textual information may be missing or incorrectly recognized, substantially increasing the difficulty of table parsing and comprehension.

To quantitatively assess the impact of these issues on model performance, we conducted an experimental evaluation using Qwen3-VL-8B~\cite{qwen3technicalreport} as a representative model. As shown in \cref{fig:sub1}, the model’s reasoning performance significantly declines when the full table (FT) is not accessible (w/o FT vs w/ FT). Additionally, performance also deteriorates in more structurally complex table formats, such as three-line tables. These observations indicate that relying solely on visual inputs for table understanding exhibits notable limitations in complex and incomplete scenarios.

\textbf{\underline{Challenge 2}: Implicit and complex feature dependencies.} Current language models struggle to identify the feature dependencies that are relevant for prediction tasks. In TVMU tasks, models typically need to capture the dependencies between features and the prediction targets. This process involves not only the direct influence of individual cells or attributes on the target but also the interrelationships among features. For example, in a table prediction task concerning pet popularity~\cite{petfinder-pawpularity-score} (\cref{fig:sub2} using Qwen3-VL-8B~\cite{qwen3technicalreport} as an example), the model may analyze only the pet images and the corresponding features in the table. However, it fails to recognize the specific contributions of each feature to popularity and their interdependencies. This limitation results in significantly degraded performance.

\textbf{\underline{Challenge 3:} The significant heterogeneity of task requirements.} 
TVMU tasks exhibit significant heterogeneity across different task settings, leading to substantially different problem-solving pipelines. These tasks differ in their reliance on information sources and reasoning processes. For example, question answering often requires integrating tabular, textual, and visual information, whereas classification tasks primarily depend on the internal structure and explicit features of tables. In addition, variations in data preprocessing further exacerbate this heterogeneity. 

Moreover, task heterogeneity poses additional challenges to model generalization: models must not only capture complex feature dependencies within and across tables, but also adapt to diverse reasoning logics and information integration strategies. Such multi-level differences complicate the design of unified models and make it difficult for a existing approach to maintain robust performance across diverse tasks.
\begin{figure*}[t]
    \centering
    \includegraphics[width=0.9\linewidth]{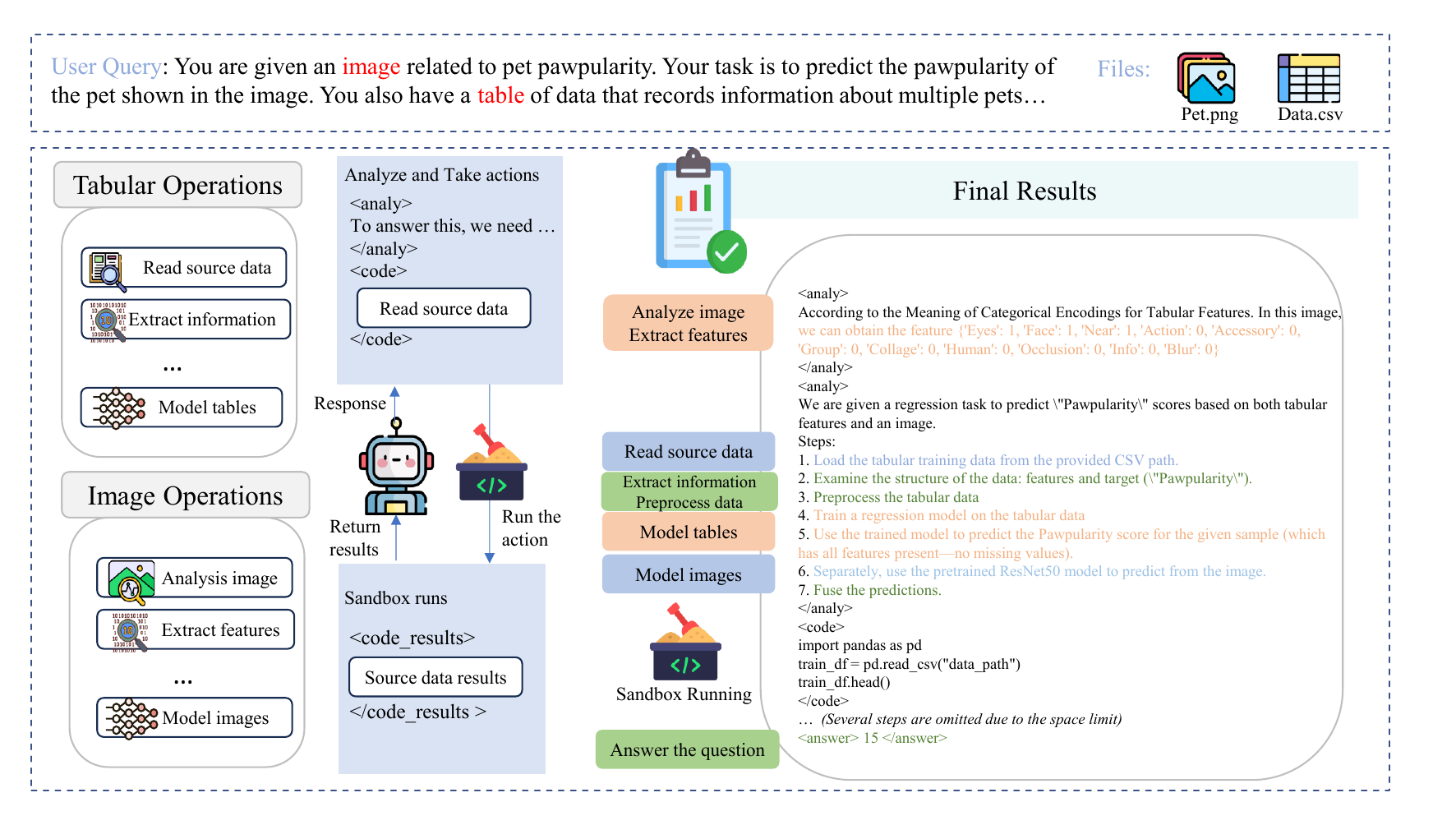}
    \caption{TWT framework. It realizes robust and good performance via code-base neuro-symbolic reasoning in TVMU tasks.}
    \label{fig:main_figure}
\end{figure*}
\section{Methods}
Inspired by the challenge above, we propose TWT, a \textbf{program-aided neuro-symbolic reasoning framework} for TVMU tasks. We design a two-stage training strategy for TWT to address the three challenges in TVMU tasks. 
In the first stage, Task-Oriented Supervised Fine-Tuning guides the model to master the processing pipelines of different table understanding tasks, thereby mitigating task heterogeneity, while enabling the model to initially learn how to parse complex table formats and model feature dependencies. In the second stage, Adaptive Loss-Scaled Reinforcement Learning is introduced to further enhance the model’s ability to handle diverse table structures and complex feature dependencies, as well as to improve the correctness and stability of its outputs. As shown in \cref{fig:main_figure}, under this training framework, TWT integrates code execution with neuro-symbolic reasoning, enabling explicit step-by-step planning and execution, while supporting flexible processing of multimodal table data and efficient information extraction, computation, and reasoning.
\subsection{Environment Setup}
\textbf{Interaction Pattern.}
To facilitate systematic problem analysis, interaction with external environments, acquisition of execution feedback, and generation of final answers, we design a unified and strictly constrained interaction protocol, which the model is required to follow for output. The protocol consists of the following components:
\begin{itemize}
    \item \texttt{<analy></analy>}: explicitly represents the model's reasoning process for the given problem;
    \item \texttt{<code></code>}: specifies the code to be executed within a sandboxed environment;
    \item \texttt{<code\_results></code\_results>}: returns the execution results or error messages produced by the sandbox;
    \item \texttt{<answer></answer>}: outputs the final answer.
\end{itemize}

In practical applications, this interaction protocol enables accurate extraction of the model's intent via simple regular expression matching and allows safe execution of generated code in a controlled sandbox environment, thereby closing the loop between reasoning, execution, and feedback and improving the system's usability and reliability.

\textbf{Sandbox Environment.}
To ensure that the model can efficiently and safely interact with files in the external environment via code, we construct a controlled sandbox environment specifically designed to execute model-generated code. This environment imposes strict constraints on code execution and provides the model with complete and accurate feedback on both outputs and any errors encountered during execution.

In practice, we observe that insufficiently trained models may generate faulty code, such as scripts containing infinite loops or unexpected behavior. To mitigate this, the sandbox environment executes code within a Python REPL equipped with a timeout mechanism, effectively preventing unbounded execution. Additionally, the sandbox pre-integrates all task-relevant resources, including data files, images, and pretrained models. Within this environment, the model can flexibly access and manipulate these resources according to specified paths, with execution results returned for subsequent reasoning and decision-making.

\subsection{Task-Oriented Supervised Fine-Tuning}
As discussed earlier, there exists a significant challenge of heterogeneity between different task types in TVMU tasks. Therefore, to construct a unified task-solving framework, the model must not only comprehend the table content itself but also possess the capability to adapt to and execute the pipelines corresponding to different tasks.

To achieve this, we first perform Task-Oriented Supervised Fine-Tuning (TO-SFT) to preliminarily train the model. This stage guides the model to effectively internalize the solution paradigms of different tasks as part of the model’s capability.

\textbf{Data Synthesis Process.} To enable this, we first need to construct a set of high-quality training data that explicitly contains problem-solving trajectories. Existing TVMU datasets, such as WikiTQ~\cite{pasupat2015compositional} and Adoption~\cite{petfinder-adoption-prediction}, while large in scale, typically provide only final answers and lack corresponding reasoning process annotations. This limitation restricts the model’s ability to learn and generalize reasoning skills. To address this, we design a verifiable data synthesis pipeline to automatically construct training samples with reliable reasoning traces, as follows.

Specifically, for the $i$-th sample, we submit its visual input ($V_i$), problem description ($Q_i$), and external environment ($E_i$) to a more capable multimodal large language model (Qwen3-VL-Plus~\cite{qwen3technicalreport}) according to a predefined input format and problem-solving prompts.

\[
x_i = \left( V_i, Q_i, E_i \right),
\]
\[
 \left( \hat{y}_i, \hat{r}_i \right) 
= \mathcal{M}(x_i),
\]
During this process, the model is required to explicitly output both its reasoning trajectory $\hat{r}_i$ and the final answer $\hat{y}_i$. The generated answers are then compared with the ground-truth annotations. For regression answers, a relative error of up to 10\% is tolerated; samples satisfying this criterion are accepted, while those exceeding it are rejected.

\[
\mathcal{A}(\hat{y}_i, y_i) =
\begin{cases}
\mathbb{I}\Biggl( \dfrac{|\hat{y}_i - y_i|}{\max(|y_i|, \epsilon)} \le \delta \Biggr), & i \in \mathcal{R},\\[1ex]
\mathbb{I}(\hat{y}_i = y_i), & i \in \mathcal{C},
\end{cases}
\]
\[
\mathcal{D}_{\text{accept}} =
\left\{ 
\left( x_i, \hat{r}_i, \hat{y}_i \right)
\;\middle|\; 
\mathcal{A}\big(\hat{y}_i, y_i\big) = 1
\right\}.
\]
where $\mathcal{R}$ denotes the set of indices for the regression task in table prediction, and  $\mathcal{C}$ denotes the set of indices for others.

However, in practical applications, even state-of-the-art MLLMs exhibit notable limitations when performing advanced operations such as table modeling and precise numerical computation. Given that large language models (LLMs)~\cite{guo2025deepseek} demonstrate more stable performance in instruction following and complex operations, we adopt a data synthesis strategy that combines MLLMs and LLMs. Specifically, the MLLM is primarily responsible for semantically understanding the input images and generating intermediate steps and structured representations for problem-solving. These outputs, along with subsequent reasoning steps, are then provided as input to the LLM, which performs more refined reasoning and computation to produce the final answer.

\textbf{Training Process}. The core objectives of this training process include internalizing problem-solving pipelines corresponding to different task types, which reduces the heterogeneity influence, and learning how to interpret and utilize results or error messages returned from code execution, thereby supporting reasoning based on execution feedback, which is helpful to process diverse table structures, incomplete data, and complex feature dependencies.

To achieve this, during the TO-SFT stage, we adopt a single-turn dialogue training setup. Specifically, for each sample in the synthesized dataset, the multi-turn reasoning path $r$ are concatenated to form a complete training sequence. 

During training, to prevent the model from directly memorizing code execution results during the TO-SFT stage~\cite{zhang2025deepanalyze,zhang2025thyme}, we mask the outputs of code execution, computing the loss only over the model-generated text and code sequences. Accordingly, the TO-SFT stage optimization objective is defined as:

\begin{equation}
    \mathcal{L}_{\text{TO-SFT}} 
= - \mathbb{E}_{(x_i, y_i) \sim \mathcal{D}_{\text{accept}}} 
\sum_{t \in \mathcal{T}_{\text{gen}}} 
\log p_\theta \big( y_{i,t} \mid x_i, y_{i,<t} \big)
\end{equation}

where $\mathcal{T}_{\text{gen}}$ denotes the set of generated tokens that are not masked.

\subsection{Adaptive Loss-Scaled Reinforcement Learning}
After completing TO-SFT training, the model has largely mastered the prescribed output conventions and basic problem-solving procedures. To further enhance the model’s ability to process TVMU tasks, we apply Adaptive Loss-Scaled Group Relative Policy Optimization (AL-GRPO) for reinforcement learning. Unlike supervised learning, AL-GRPO optimizes the model via policy gradients derived from reward signals, minimizing token-level prediction errors while encouraging the model to explore a broader solution space. This facilitates learning more diverse and complex reasoning behaviors.

In prior work~\cite{guo2025deepseek}, GRPO reward design often emphasizes the format correctness of model outputs. However, we observe that after high-quality TO-SFT training, the model can reliably generate responses according to predefined formats, making format constraints no longer a primary bottleneck. Based on this observation, during the reinforcement learning stage, we focus the reward function on the correctness of the final answer, which can be formalized as follows:

\[
\text{Reward}(\hat{y}_i, y_i) =
\begin{cases}
\displaystyle
1 - \min\!\left(
\dfrac{|\hat{y}_i - y_i|}{|y_i| + \epsilon},\; 1
\right), & i \in \mathcal{R},\\[2ex]
\mathbb{I}(\hat{y}_i = y_i), & i \in \mathcal{C}.
\end{cases}
\]

This reward formulation combines both binary (0/1) signals and continuous-valued feedback, effectively alleviating the issue of reward sparsity.

While prioritizing the final answer correctness, we also aim for the model to generate executable and stable code, reducing invalid interactions with the external environment. Directly using code correctness as a reward is challenging: (1) code complexity varies significantly across samples, and (2) rewarding only based on successful execution leads to extremely sparse rewards for code involving complex modeling or multi-step computation, which destabilizes training.

To address this, we introduce an adaptive loss-scaled strategy to determine which code generation steps contribute to the AL-GRPO loss. Only code segments that terminate successfully are included in the AL-GRPO loss computation, while segments that fail or terminate abnormally receive no gradient updates. By excluding code-related information from the reward, this approach both prevents reward hacking and allows the model to gradually favor executable and stable code without requiring complex code-level reward definitions.

For each question \( x_i \) in the training dataset \( \mathcal{D} \), AL-GRPO samples a group of \( G \) multi-turn outputs
\( \{\boldsymbol{o}_1, \ldots, \boldsymbol{o}_G\} \) from the old policy \( \pi_{\text{old}} \).
Consider the code execution situation, we obtain $\boldsymbol{\hat{o}}$, which satisfies.
\[
\boldsymbol{\hat{o}}_i = (o_i^1, o_i^2, \ldots, o_i^{n_i}), \quad \text{where } S(o_i^k)=1 \ \forall k.
\]
Here $S(\cdot)$ is the sandbox output about whether the code exited normally.
The policy model \( \pi_\theta \) is then optimized by maximizing the following objective:
\begin{equation}
\begin{split}
\mathcal{L}_{\text{AL-GRPO}} =
 \; &\mathbb{E}_{q \sim \mathcal{D},\, \{\boldsymbol{\hat{o}}_i\}_{i=1}^{G} \sim \pi_{\theta_{\text{old}}}(\cdot \mid q)} \Biggl[ \frac{1}{G} \sum_{i=1}^{G} \Biggl(
\min \\ \Biggl(
\frac{\pi_\theta(\boldsymbol{\hat{o}}_i \mid q)}{\pi_{\theta_{\text{old}}}(\boldsymbol{\hat{o}}_i \mid q)} A_i, 
&  \mathrm{clip}\Bigl(
\frac{\pi_\theta(\boldsymbol{\hat{o}}_i \mid q)}{\pi_{\theta_{\text{old}}}(\boldsymbol{\hat{o}}_i \mid q)}, 
1 - \epsilon, 
1 + \epsilon
\Bigr) A_i
\Biggr) \\
& \quad\quad\quad\quad\quad\quad - \beta \, D_{\mathrm{KL}}\!\left(
\pi_\theta \,\|\, \pi_{\text{ref}}
\right)
\Biggr)
\Biggr]
\end{split}
\end{equation}

where \(A_i\) is the advantage calculated from the rewards \(\{r_1, \dots, r_G\}\) of outputs within each group, \(\pi_{\text{ref}}\) is the reference model, and \(\epsilon\) and \(\beta\) are hyperparameters.

\begin{table*}[t]
\caption{Comparison of different models on multimodal table question answering and multimodal table prediction benchmarks. The best results are shown in \textbf{bold}, and the second-best results are underlined.}

\centering
\resizebox{0.95\textwidth}{!}{
\begin{tabular}{lccccccccc}
\toprule
\multirow{2}{*}{\textbf{Models}} 
& \multicolumn{4}{c}{\textbf{Multimodal Table QA}} 
& \multicolumn{4}{c}{\textbf{Multimodal Table Prediction}} 
& \textbf{AVG\_rank} \\
\cmidrule(lr){2-5} \cmidrule(lr){6-9}

& \textbf{TabMWP} 
& \textbf{WikiTQ} 
& \textbf{FinQA} 
& \textbf{TAT-QA} 
& \textbf{Adoption} 
& \textbf{SkinCA} 
& \textbf{Pawpularity} 
& \textbf{Paintings} 
&  \\

& \small ACC $\uparrow$ 
& \small ACC $\uparrow$ 
& \small ACC $\uparrow$ 
& \small ACC $\uparrow$ 
& \small ACC $\uparrow$ 
& \small ACC $\uparrow$ 
& \small nRMSE $\downarrow$ 
& \small nRMSE $\downarrow$ 
& - \\
\midrule

\multicolumn{10}{c}{\textbf{API-based MLLMs}} \\
\midrule
\textbf{GPT-5.1}      & 92.50 & 40.82 & 47.68 & 56.38 & \underline{26.12} & \underline{65.51} & \underline{0.366} & \underline{0.098} & 3.63 \\
\textbf{Qwen3VL-Plus}    & \textbf{98.70} & \underline{76.10} & \textbf{64.41} & \underline{72.39}  & 18.67 & 51.55 & 0.539 & 0.178 & 4.25 \\

\midrule
\multicolumn{10}{c}{\textbf{ General Open-Source MLLMs}} \\
\midrule
\textbf{Qwen3VL-8B}        & 86.70 & 50.64 & 38.24 & 62.53  & 25.37 & 40.38 & 0.501 & 0.172 & 5.63 \\
\textbf{Qwen3VL-30B}       & 88.10 & 58.35 & 55.76 & 59.17  & 18.74 & 44.67 & 0.493 & 0.130 & 5.00 \\
\textbf{InternVL3.5-8B}    & 82.80 & 47.97 & 42.21 & 54.87  & 19.41 & 43.20 & 0.466 & 0.119 & 5.75 \\
\textbf{InternVL3.5-38B}   & 90.10 & 67.84 & 57.77 & 60.37  & 20.52 & 43.47 & 0.495 & 0.110 & 4.38 \\
\midrule
\multicolumn{10}{c}{\textbf{Tabular Related Open-Source MLLMs}} \\
\midrule
\textbf{HIPPO}             & 79.90 & 30.68 & 7.86 & 18.69 & 23.68 & 29.76 & 0.485 & 0.290 & 7.25 \\
\textbf{Table-LLAVA}       & 30.15 & 9.16 & 4.50 & 11.75  & 23.19 & 30.28 & 0.494 & 0.282 & 7.88 \\

\midrule
\rowcolor[gray]{.95}
\textbf{TWT (Ours)} 
& \underline{97.10} 
& \textbf{78.95} 
& \underline{61.95} 
& \textbf{72.82}  
& \textbf{32.90} 
& \textbf{73.65} 
& \textbf{0.199} 
& \textbf{0.093} 
& \textbf{1.25} \\

\bottomrule
\end{tabular}
}
\label{tab:mainres}
\end{table*}

\section{Experiments}
In this section, we first describe the overall experimental setup, including the datasets, evaluation metrics, and implementation details. We then report the main experimental results, accompanied by both quantitative and qualitative analyses. Finally, we summarize the training process and further assess the model’s robustness across different scenarios.
\subsection{Experimental Setup}
\textbf{Implementation Details}. We adopt Qwen3-VL-8B~\cite{qwen3technicalreport} as the base model for TWT. During the data synthesis stage, approximately 1.5K high-quality multimodal question–answering training samples are generated with the assistance of Qwen3VL-Plus and Qwen3-Max~\cite{qwen3technicalreport}. For multimodal table prediction tasks, we collect data from four datasets, resulting in a total of 1.2K synthesized training samples. In the TO-SFT stage, the model is trained for three epochs on a 4× A800 server with a learning rate of 1e-5. In the AL-GRPO stage, we select 0.5K QA samples and 0.4K prediction samples from the synthesized data to perform training on an 8×A800 server for two epochs, using a learning rate of 5e-7.

\textbf{Evaluation Datasets and Metrics.} We evaluate model performance primarily on two types of tasks: multimodal table question answering and table prediction. For QA tasks, we adopt four benchmark datasets: WikiTQ~\cite{pasupat2015compositional}, TabMWP~\cite{lu2022dynamic}, FinQA~\cite{chen2021finqa}, and TAT-QA~\cite{zhu2021tat}. Among them, WTQ and TabMWP are standard table question answering datasets, while FinQA and TAT-QA additionally include table-related contextual text for each question. For table prediction, we evaluate on four multimodal tabular datasets: Adoption~\cite{petfinder-adoption-prediction} and SkinCA\footnote{https://www.kaggle.com/datasets/mahdavi1202/skin-cancer} for classification tasks, and Pawpularity~\cite{petfinder-pawpularity-score} and Paintings\footnote{https://www.kaggle.com/datasets/denozavrus/paintings-price-prediction/data} for regression tasks. In terms of evaluation metrics, we use accuracy (ACC) for question answering and classification tasks, and normalized Root Mean Squared Error (nRMSE) for regression tasks. Notably, during evaluation, we do not directly provide the model with complete table contents. Instead, only the table headers are included in the visual input, and the full table data must be retrieved by the model through code-based interaction with the external environment. This evaluation setting more closely reflects real-world application scenarios and enables a more realistic assessment of the model’s capabilities in complex multimodal table understanding tasks.

\textbf{Comparison Methods.} In our comparative experiments, we select a diverse set of representative models covering both mainstream open-source multimodal large language models and tabular understanding approaches. Specifically, the evaluated open-source models include Qwen3-VL-8B / 30B~\cite{qwen3technicalreport}, InternVL-3.5-8B / 38B~\cite{wang2025internvl3}, HIPPO~\cite{liu2025hippo}, and Table-LLaVA~\cite{zheng2024multimodal}. In addition, we include two proprietary API-based models as reference upper bounds: GPT-5.1 and Qwen3-VL-Plus. All baselines are provided with identical sandbox instructions and guaranteed access to its resources.

\subsection{Main Results}
In our experiments, we evaluated TWT and the baseline models across eight TVMU datasets. As shown in \cref{tab:mainres}, TWT consistently outperforms open-source models and achieves the best overall performance, demonstrating clear and stable improvements over existing baselines. 

Specifically, on QA tasks, which mainly assess the ability to handle diverse table structures and incomplete data, TWT achieves nearly a 10\% improvement over the strongest baseline and delivers performance comparable to API-based models. This advantage primarily stems from TWT’s training procedure, which enables the model to effectively understand diverse table structures and perform robust information extraction. In contrast, baseline models such as HIPPO~\cite{liu2025hippo} exhibit weaker performance: although they are trained for table understanding, they have not been exposed to scenarios with missing or incomplete information, leading to inferior robustness and degraded performance in TVMU settings.

For table prediction tasks, TWT also shows significantly higher performance compared to the baselines. This is because TWT models feature-target relationships through code execution, which captures complex dependencies more effectively than the analytical modeling methods used by the baselines. Additionally, TWT leverages pretrained models for the image modality, further improving prediction accuracy—an ability that baseline models lack.

Overall, the strong performance of TWT on both tasks demonstrates its effectiveness in handling heterogeneous tasks. This effectiveness can be attributed to the internalization of task-solving pipelines during training. Experimental results further show that, through the proposed two-stage training strategy, TWT is able to effectively address the three aforementioned challenges.
\subsection{Further Analysis}

\textbf{Ablation Study}
To better investigate the contributions of the training process to TWT, we conducted ablation experiments on four datasets, as shown in \cref{tab:abla}. It can be observed that, after the TO-SFT stage, the model has already learned the basic problem-solving procedures for different types of TVMU tasks, acquiring accurate analytical steps and generating correct code to interact with the environment to answer questions. Following AL-GRPO, the model shows further improvements on more complex problems and demonstrates a stronger understanding of both visual and textual modalities, resulting in overall enhanced performance. This indicates that the two stages of our framework are complementary and jointly essential for maximizing the model’s reasoning capabilities.
\begin{table}[htbp]
\centering
\small
\caption{Experimental Results of Qwen3-VL-8B under TO-SFT and AL-GRPO on Table QA and Table Prediction Benchmarks. Best is shown in \textbf{bold}}
\resizebox{0.95\linewidth}{!}{
\begin{tabular}{lcccc}
\toprule
\multirow{2}{*}{\textbf{Method}} 
& \multicolumn{2}{c}{\textbf{Table QA}}  
& \multicolumn{2}{c}{\textbf{Table Prediction}}  \\
\cmidrule(lr){2-3} \cmidrule(lr){4-5}

& \textbf{WikiTQ} 
& \textbf{TAT-QA} 
& \textbf{SkinCA}  
& \textbf{Paintings}  \\

& \small ACC $\uparrow$ 
& \small ACC $\uparrow$ 
& \small ACC $\uparrow$ 
& \small nRMSE $\downarrow$  \\
\midrule
\textbf{Qwen3VL-8B}   &50.64 & 62.53 & 40.38 & 0.172 \\
\textbf{+TO-SFT}  & 78.38 & 71.31 & 71.08 & 0.099  \\
\textbf{+AL-GRPO}   & \textbf{78.95} & \textbf{72.82} & \textbf{73.65} & \textbf{0.093} \\
\bottomrule
\end{tabular}
}
\label{tab:abla}

\end{table}

\textbf{Performance on Situations without tabular features.}
To further evaluate TWT’s performance under incomplete information, we deliberately removed the table content from the sample images in the multimodal table prediction. When the table features in the images are missing, TWT cannot directly rely on complete table modeling and must instead extract relevant features from the images to accomplish the task. As shown in \cref{tab:miss}, even under incomplete information, TWT maintains competitive performance. This demonstrates that, in multimodal table prediction scenarios, TWT can effectively align table and visual information based on image understanding, handle missing values, and achieve robust predictions.

\begin{table}[htbp]
\centering
\small
\caption{Experiment results of TWT under situation w and w/o table features in multimodal table prediction tasks}
\begin{tabular}{lcc}
\toprule
\multirow{2}{*}{\textbf{Method}} 
& \textbf{Adoption}  
& \textbf{Pawpularity}  \\
\cmidrule(lr){2-2} \cmidrule(lr){3-3}
& \small ACC $\uparrow$ 
& \small nRMSE $\downarrow$  \\
\midrule
\textbf{TWT w features}   & 32.90 & 0.199 \\
\textbf{TWT w/o features}  & 31.40 & 0.200 \\
\bottomrule
\end{tabular}
\label{tab:miss}
\end{table}

\textbf{Code Execution Accuracy.}
\begin{figure}[t]
    \centering
    \includegraphics[width=0.7\linewidth]{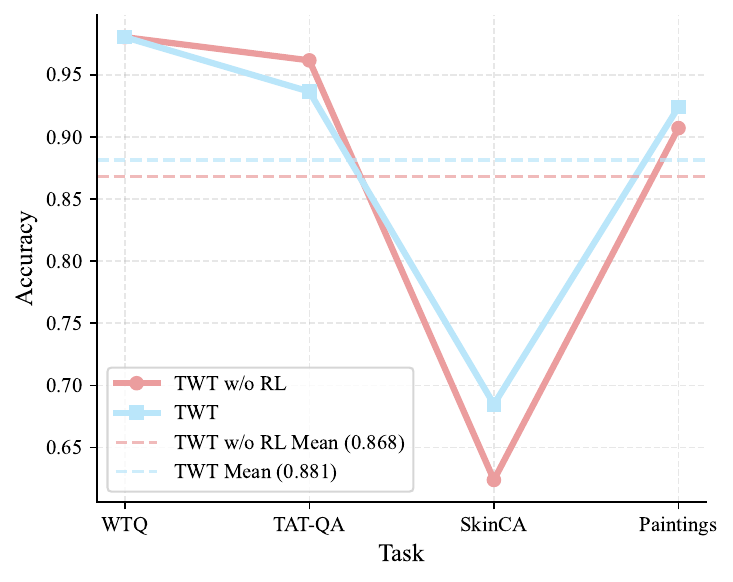}
    \caption{Comparison of code execution accuracy of TWT w and w/o AL-GRPO}
    \label{fig:accucode}
\end{figure}
During the RL training stage, we adopt a dynamic loss scaling strategy to improve the stability and accuracy of the generated code. To evaluate its effectiveness, we measure the code execution accuracy during the problem-solving process across multiple datasets, using the same datasets as in the ablation study. The results are shown in \cref{fig:accucode}. As can be observed, after AL-GRPO training, TWT achieves higher code execution accuracy on most tasks.

On the TAT-QA dataset, a slight decrease in code execution accuracy is observed. However, when considered together with the overall performance improvements on this dataset, the results indicate that the model’s reasoning capability on complex tasks is still significantly enhanced after AL-GRPO training. This suggests that dynamic loss scaling helps guide the model toward more effective reasoning and decision-making processes; even when code execution accuracy does not consistently increase, it still contributes to improved overall reasoning performance and final task accuracy.

\section{Conclusion}
In this paper, we study the problem of TVMU tasks and, through an in-depth analysis, identify three fundamental challenges inherent to this task. To address these challenges, we propose TWT, \textbf{a program-aided neuro-symbolic reasoning} framework that equips models with code-based reasoning capabilities through a two-stage training strategy. By learning to plan and execute key problem-solving steps via effective interaction with external code execution environments, TWT enables reliable reasoning over multimodal tabular data. Experiments on eight datasets demonstrate the effectiveness of our approach.

\section*{Impact Statement}

This work aims to advance machine learning in tabular–vision multimodal understanding, enhancing models’ reasoning capabilities over complex multimodal data. We do not foresee any significant ethical risks or negative societal impacts from our method, but we encourage future research to consider responsible use and potential impacts in real-world applications.

\bibliography{example_paper}
\bibliographystyle{icml2026}

\newpage
\appendix
\onecolumn
\section{Detailed Experiment Setup}

\subsection{Dataset Detail}
In this section, we introduce the 8 datasets used in our experiments.

For multimodal table question answer dataset.
\begin{enumerate}
    \item WikiTQ~\cite{pasupat2015compositional} A table-based question answering dataset focused on general knowledge, emphasizing compositional reasoning.
    \item TabMWP~\cite{lu2022dynamic} A table-based math word problem dataset requiring dynamic calculation and reasoning over tables.
    \item FinQA~\cite{chen2021finqa} A financial table QA dataset that evaluates models’ numerical understanding and computation abilities on financial tables.
    \item TAT-QA~\cite{zhu2021tat} A hybrid text–table QA dataset that requires complex reasoning by combining textual information with table content.
\end{enumerate}

For multimodal table prediction dataset.
\begin{enumerate}
    \item Adoption~\cite{petfinder-adoption-prediction} Classification task. Predict the speed at which a pet is adopted, based on the pet’s listing on PetFinder. 
    \item SkinCA\footnote{https://www.kaggle.com/datasets/mahdavi1202/skin-cancer} Classification task. This dataset provides a diverse collection of skin lesion images, covering both malignant and benign types, with associated metadata, intended to support research in skin cancer detection and dermatological analysis.
    \item Pawpularity~\cite{petfinder-pawpularity-score} Regression task. This dataset contains pet profile images and associated metadata, with engagement measured by the Pawpularity Score, which reflects normalized page views while excluding duplicate clicks, bots, and sponsored profiles.
    \item Paintings\footnote{https://www.kaggle.com/datasets/denozavrus/paintings-price-prediction/data} Regression task. The goal of this task is to predict the price of paintings based on their images and tabular features.
\end{enumerate}

\subsection{Implementation Details}
In TO-SFT stage, we use a dataset consisting of 1.5K QA data and 1.2K prediction comes from our data synthesis process. Additionally, about 3.0K data comes from math dataset and code dataset\footnote{https://huggingface.co/datasets/RUC-DataLab/DataScience-Instruct-500K/tree/main/reasoning}. The used dataset will be open sourced.

In this experiment, we perform full fine-tuning of the Qwen3-vl-8b model using 4 A800-80G GPUs. Model parameters are trained in bfloat16 precision, with the visual encoder frozen. Training is conducted for 3 epochs, with a per-GPU batch size of 1 and gradient accumulation to achieve an effective batch size of 16. The maximum input length is set to 15,360 tokens, the learning rate is 1e-5, and the warmup ratio is 0.05. Attention computation is accelerated using Zero3 optimization.

In AL-GRPO stage,  we use a dataset consisting of 0.5K QA data and 0.4K prediction comes from our data synthesis process.

We perform AL-GRPO using 8 A800-80G GPUs. The training is conducted for 2 epochs with a per-GPU batch size of 1, and gradient accumulation over 4 steps to achieve a larger effective batch size. The maximum input and output sequence lengths are set to 10,240 tokens, with a maximum pixel count of 2,408,448. The learning rate is 5e-7, scheduled using a cosine scheduler with a minimum learning rate, and a warmup ratio of 0.03. Training is performed using Zero3 optimization.

\newpage
\section{Prompt Used for Data Synthesis}

\begin{formalizedprompt}[title=Prompt for QA data synthesis]

\begin{quote}
In a table QA task you need to answer the following question: \{\}, The image shows a partial sample of the table (df.head()). The image is only to help you understand the table's format. You must, from understanding the table format shown in the image, analyze the table's rows and columns and write Python code to interact with the entire table in order to answer the question. Any information required to answer the question must be obtained from the table; when extracting information, be robust to issues such as case differences, abbreviations, etc. Do not use your background knowledge. Wrap your analysis in \textless analy\textgreater\textless/analy\textgreater. Wrap code that needs to run in \textless code\textgreater\textless/code\textgreater. The interactions will run in a sandbox and the results will be provided to you — use those results for reasoning. Perform one operation at a time until you can answer the question, and wrap the final answer in \textless answer\textgreater\textless/answer\textgreater.

The CSV file path is: \{\},

Your previous operation paths and outputs were: \{\}.
\end{quote}
\end{formalizedprompt}
\newpage
\begin{formalizedprompt}[title=Prompt for prediction data synthesis for MLLMs]
\begin{quote}
I will give you a image, you need to make predictions for this image. Task describe and instructions are as follow:

\#\# Task is \{\}

You need to extract the features I want from this image; if you cannot extract them, leave them blank.

\{ Feature descriptions\}

\#\#\# Output Formate
\textless  answer\textgreater JSON Formate \textless /answer\textgreater

\end{quote}
\end{formalizedprompt}

\begin{formalizedprompt}[title=Prompt for prediction data synthesis for LLMs]
\begin{quote}
I will give you a sample whose features are extracted from an image. The sample’s features correspond to those in a tabular file. I now need you to build a model based on the tabular data and then make a prediction for the sample I provide. I will give you some hints.

Solve the following table problem step by step (one code block at a time), and optionally write Python code to enhance your reasoning process. The Python code will be executed by an external sandbox, and the processed result (wrapped in \textless code\_result\textgreater\textless /code\_result\textgreater) can be returned to aid your reasoning and help you arrive at the final answer.

\#\# task

\{\}

\#\# instructions

1. You need to first analyze the tabular data and the sample, then build a predictive model and then If you can, optimize the model performance and then evaluate it.

2. Since the provided sample is obtained from an image, some features may be missing. During prediction, please properly handle missing values; do not rely only on the shared features or leave them as NaN.

3. This is a multimodal prediction task, so we provide a pretrained ResNet50 model. You can call it as follows:
   `predict = ResNet50Predict(task, image\_path)`
   
4. Report the tabular prediction(in predict\_proba format) and image prediction in the final answer with format (tabular\_prediction, image\_prediction).

\#\#\# Files
sample: \{\}

tabular data path: \{\}

image path is: \{\}

\textless analy\textgreater Your detailed reasoning process should go here.\textless /analy\textgreater

\textless code\textgreater(Optional) the code that need to run in the sandbox\textless /code\textgreater

\textless answer\textgreater Your final answer to the user's question goes here.\textless /answer\textgreater

\textless code\_result\textgreater\textless /code\_result\textgreater is obtained from a external sandbox, you must not generate it by your own.
'''

\end{quote}
\end{formalizedprompt}

\newpage

\section{Example for TWT Replies}
\begin{figure}[h]
    \centering
    \includegraphics[width=\linewidth]{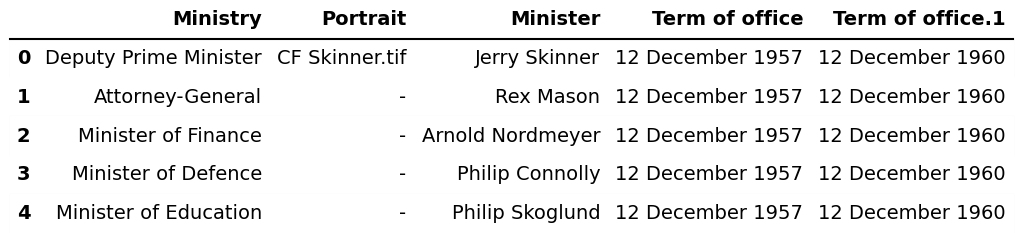}
    \caption{\textless image \textgreater }
\end{figure}
\begin{formalizedprompt}[title=Replies for QA tasks]

\begin{quote}

\textbf{Queries:}

\textless image\textgreater 

\#\#\# Question

Who served as the Minister of Foreign Affairs for Second Labour Government of New Zealand, and what dates did he serve?

\#\#\# Table file path: xxx

\textbf{Responses of TWT}

\textless analy\textgreater 

The user is asking for the Minister of Foreign Affairs during the Second Labour Government of New Zealand and their term dates. The sample table shows ministries like Deputy Prime Minister, Attorney-General, etc., but does not include \"Minister of Foreign Affairs\". To answer this, I need to read the full CSV file and search for the relevant ministry entry.

First, I'll load the CSV file and check the actual column names and data. The sample suggests columns like "Ministry", "Minister", and "Term of office", but the exact structure must be verified. I'll print the first few rows to understand the data format and check for possible variations in ministry titles (e.g., "Minister for Foreign Affairs" instead of "of").

\textless /analy\textgreater 

\textless code\textgreater 

import pandas as pd

df =pd.read\_csv('xx')

print(df.head().to\_markdown(index=False))

\textless /code\textgreater 

\textless code\_result\textgreater 

(Omitted due to space)
\textless /code\_result\textgreater 

\textless analy\textgreater 

The initial sample shows the first five rows of the table, none of which mention "Minister of Foreign Affairs" or similar. To answer the question, I need to check if the full table contains a row for the Minister of Foreign Affairs (or External Affairs, as it was historically called in New Zealand). I will search the entire 'Ministry' column for entries containing "External Affairs" or "Foreign Affairs" (case-insensitive) to identify the relevant row.

\textless /analy\textgreater 

\textless code\textgreater 

import pandas as pd

df = pd.read\_csv('xx')\# Search for ministries related to Foreign/External Affairs

foreign\_affairs\_rows = df[df['Ministry'].str.contains('External Affairs\textbar Foreign Affairs', case=False, na=False)]

print(foreign\_affairs\_rows[['Ministry', 'Minister', 'Term of office', 'Term of office.1']].to\_markdown(index=False))

\textless /code\textgreater 

\textless code\_result\textgreater 

(Omitted due to space)

\textless /code\_result\textgreater 

\textless analy\textgreater 

The search revealed a row for "Minister of External Affairs" (the historical name for Foreign Affairs in New Zealand during this period). The minister was Walter Nash, serving from 12 December 1957 to 12 December 1960. This matches the term dates of the Second Labour Government as shown in the sample data.

\textless /analy\textgreater 

\textless answer\textgreater 

Walter Nash served as the Minister of External Affairs (the historical title for Foreign Affairs in New Zealand) from 12 December 1957 to 12 December 1960.

\textless /answer\textgreater

\end{quote}
\end{formalizedprompt}

\newpage

\begin{figure}[h]
    \centering
    \includegraphics[width=0.5\linewidth]{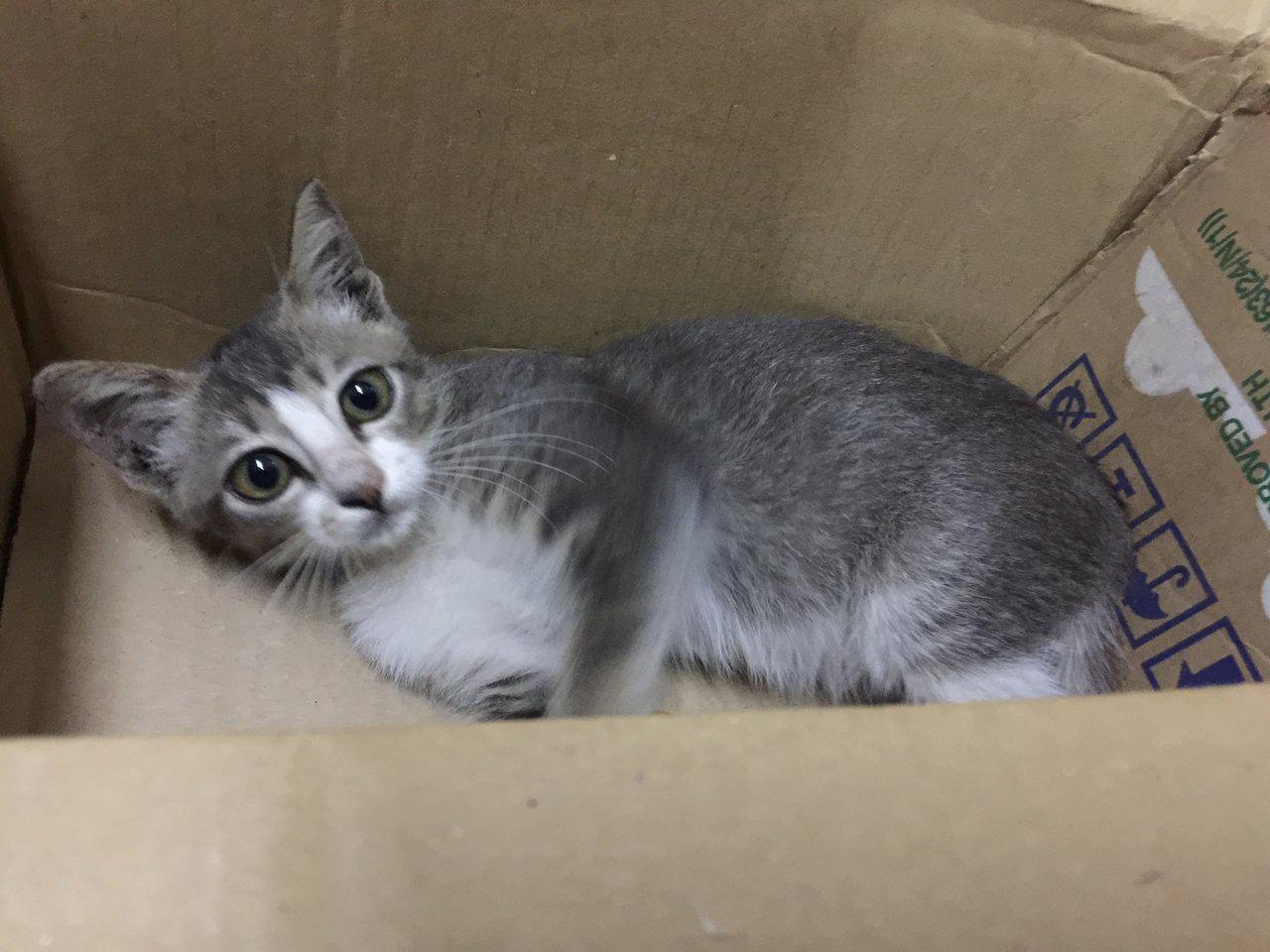}
    \caption{\textless image \textgreater }
\end{figure}
\begin{formalizedprompt}[title=Replies for QA tasks]

\begin{quote}

\textbf{Queries:}

You are given an image related to pet pawpularity. Your task is to predict the pawpularity of the pet shown in the image. You also have a table of data that records information about multiple pets and their pawpularity, as well as a ResNet50 model used to predict the pawpularity. You need to take the nature of the task into account and combine the prediction information from both modalities (tabular data and image data) to make a final prediction.

The image is \texttt{<image>}

Instructions:

1. First, based on the information in the table, you need to extract features for this sample from the image. The extracted features will be used together with the tabular information for prediction.

2. You need to build a tabular model to make a prediction for this sample.

3. At the same time, you need to incorporate the prediction from the image model. You can call the image model as follows:

\begin{verbatim}
predict = ResNet50Predict(task, image_path)
\end{verbatim}

4. Based on the task information, comprehensively consider both predictions and produce the final answer. You need to fuse the tabular prediction and image prediction, and finally output the final prediction.

Photo Metadata:

(Omitted due to space)

Files:

Image path: \texttt{xx} \\
Tabular path: \texttt{xx}

\textbf{Responses of TWT}

\texttt{<analy>}
According to the Meaning of Categorical Encodings for Tabular Features. 
In this image, we can obtain the feature \{'Eyes': 1, 'Face': 1, 'Near': 1, 'Action': 0, 
'Accessory': 0, 'Group': 0, 'Collage': 0, 'Human': 0, 'Occlusion': 0, 'Info': 0, 'Blur': 0\}
\texttt{</analy>}

\texttt{<analy>}
We are given a regression task to predict "Pawpularity" scores based on both tabular features and an image. 

The sample includes binary tabular features, and we also have access to a pretrained ResNet50 model for image-based prediction.

Steps:

1. Load the tabular training data from the provided CSV path.

2. Examine the structure of the data: features and target ("Pawpularity").

3. Preprocess the tabular data (handle missing values if any, though likely none in training).

4. Train a regression model on the tabular data (e.g., Random Forest or XGBoost for robustness).

5. Use the trained model to predict the Pawpularity score for the given sample.

6. Separately, use the pretrained ResNet50 model to predict from the image.

7. Return both predictions as (tabular\_prediction, image\_prediction).

First, let's load and inspect the training data.
\texttt{</analy>}

\texttt{<code>}

import pandas as pd

\# Load the training data

train\_df = pd.read\_csv("xx")

train\_df.head()

\texttt{</code>}

\texttt{<code\_result>}

(Omitted due to space limit)

\texttt{</code\_result>}

\texttt{<analy>}
The training data includes the following columns: 'Id', several binary features, and the target 'Pawpularity'. 

The sample we are to predict has all the binary features present, so no missing values in this case.

We note that the first column is 'Id', which corresponds to the image filename (without extension). The sample's image ID is 'xxxx', which matches the first row of the training data. 

However, we should not assume the sample is from the training set; we must build a general model.

Next steps:

- Prepare the feature matrix X (excluding 'Id' and 'Pawpularity') and target y.

- Train a regression model. We'll use Random Forest Regressor for simplicity.

- Predict on the provided sample using the trained tabular model.

- Separately, call the ResNet50Predict function for the image prediction.

Let's proceed with training the tabular model.
\texttt{</analy>}

\texttt{<code>}

from sklearn.ensemble import RandomForestRegressor

from sklearn.model\_selection import train\_test\_split

from sklearn.metrics import mean\_squared\_error

import numpy as np

\# Prepare features and target

feature\_cols = [col for col in train\_df.columns if col not in ['Id', 'Pawpularity']]

X = train\_df[feature\_cols]

y = train\_df['Pawpularity']

\# Split for validation (we'll use full data later)

X\_train, X\_val, y\_train, y\_val = train\_test\_split(X, y, test\_size=0.2, random\_state=42)

\# Train model

model = RandomForestRegressor(n\_estimators=100, random\_state=42)

model.fit(X\_train, y\_train)

\# Validate

y\_pred\_val = model.predict(X\_val)

rmse = np.sqrt(mean\_squared\_error(y\_val, y\_pred\_val))

rmse

\texttt{</code>}

\texttt{<code\_result>}
21.601680444759133
\texttt{</code\_result>}

\texttt{<analy>}

The Random Forest model achieves a validation RMSE of approximately 21.6. 

We'll retrain on the full training set to maximize performance, then predict on the provided sample.

Sample features:

\{'Eyes': 1, 'Face': 1, 'Near': 1, 'Action': 0, 'Accessory': 0, 'Group': 0, 'Collage': 0, 
'Human': 0, 'Occlusion': 0, 'Info': 0, 'Blur': 0\}

The training data includes a feature 'Subject Focus' not present in the sample. 

We'll impute it with the mode (0) to handle missing values.
\texttt{</analy>}

\texttt{<code>}

\# Retrain model on full dataset

full\_model = RandomForestRegressor(n\_estimators=100, random\_state=42)

full\_model.fit(X, y)

\# Prepare sample with imputed 'Subject Focus'

sample\_features = \{'Eyes': 1, 'Face': 1, 'Near': 1, 'Action': 0, 'Accessory': 0, 
                   'Group': 0, 'Collage': 0, 'Human': 0, 'Occlusion': 0, 'Info': 0, 'Blur': 0\}
                   
sample\_features['Subject Focus'] = 0  \# impute missing feature

\# Ensure feature order matches training data

sample\_df = pd.DataFrame([sample\_features])

sample\_X = sample\_df[feature\_cols]  \# feature\_cols includes 'Subject Focus'

\# Tabular prediction

tabular\_pred = full\_model.predict(sample\_X)[0]

\# Image prediction using pretrained ResNet50

image\_path = "xxx"

image\_pred = ResNet50Predict("pawpularity", image\_path)

(tabular\_pred, image\_pred)

\texttt{</code>}

\texttt{<code\_result>}

(np.float64(38.02495413188204), 94.54036712646484)

\texttt{</code\_result>}

\texttt{<analy>}

The (tabular\_prediction, image\_prediction) is (38.02495413188204, 94.54036712646484). 
We need to fuse them based on the task. 

The fusion strategy uses a weighted average with 0.9 for image prediction and 0.1 for tabular prediction, since Pawpularity is primarily visual.

\texttt{</analy>}

\texttt{<answer>}

88.88873581602657

\texttt{</answer>}

\end{quote}
\end{formalizedprompt}

Note that the final result is obtained through a fusion process, where the specific strategy is determined by the model based on factors such as data quality and task characteristics. For simplicity, we illustrate the most basic reasoning, stating only that “Pawpularity is primarily visual.” In more complex scenarios, the model also considers information such as the amount of missing data to produce a comprehensive final judgment.


\end{document}